# The SVM Classifier Based on the Modified Particle Swarm Optimization

Liliya Demidova
Moscow Technological Institute,
Ryazan State Radio Engineering University
Moscow, Russia

Evgeny Nikulchev
Moscow Technological Institute
Moscow, Russia

Yulia Sokolova
Ryazan State Radio Engineering University
Ryazan, Russia

*Abstract*—The problem of development of the SVM classifier based on the modified particle swarm optimization has been considered. This algorithm carries out the simultaneous search of the kernel function type, values of the kernel function parameters and value of the regularization parameter for the SVM classifier. Such SVM classifier provides the high quality of data classification. The idea of particles' «regeneration» is put on the basis of the modified particle swarm optimization algorithm. At the realization of this idea, some particles change their kernel function type to the one which corresponds to the particle with the best value of the classification accuracy. The offered particle swarm optimization algorithm allows reducing the time expenditures for development of the SVM classifier. The results of experimental studies confirm the efficiency of this algorithm.

## I. INTRODUCTION

Currently, for the different classification problems in various applications the SVM algorithm (Support Vector Machines, SVM), which carries out training on precedents («supervised learning»), is successfully used. This algorithm includes in the group of boundary classification algorithms [1], [2].

The SVM classifiers by the SVM algorithm have been applied for credit risk analysis [3], medical diagnostics [4], handwritten character recognition [5], text categorization [6], information extraction [7], pedestrian detection [8], face detection [9], etc.

The main feature of the SVM classifier is using of the special function called the kernel, with which the experimental data set has been converted from the original space of characteristics into the higher dimension space with the construction of a hyperplane that separates classes. A herewith two parallel hyperplanes must be constructed on both sides of the separating hyperplane. These hyperplanes define borders of classes and have been situated at the maximal possible distance from each other. It has been assumed that the bigger distance between these parallel hyperplanes gives the better accuracy of the SVM classifier. Vectors of the classified objects' characteristics which are the nearest to the parallel hyperplanes are called support vectors. An example of the separating hyperplane building in the 2D space has been shown in Fig. 1.

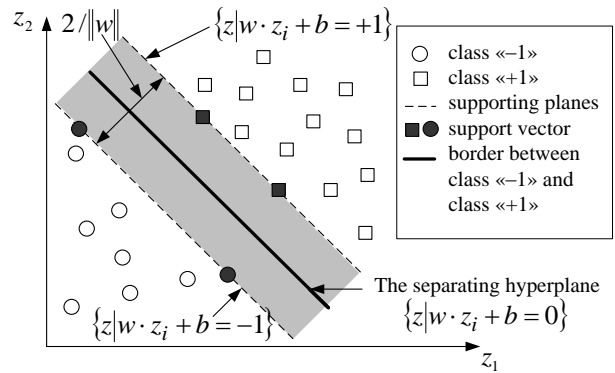

Fig. 1. Linear separation for two classes by the SVM classifier in the 2D space

The SVM classifier supposes an execution of training, testing, and classification. Satisfactory quality of training and testing allows using the resulting SVM classifier in the classification of new objects.

Training of the SVM classifier assumes solving a quadratic optimization problem [1]–[3]. Using a standard quadratic problem solver for training the SVM classifier would involve solving a big quadratic programming problem even for a moderate sized data set. This can limit the size of problems which can be solved with the application of the SVM classifier. Nowdays methods like SMO [10, 11], chunking [12] and simple SVM [13], Pegasos [14] exist that iteratively compute the required solution and have a linear space complexity [15].

A solution for the problem which has been connected with a choice of the optimal parameters' values of the SVM classifier represents essential interest. It is necessary to find the kernel function type, values of the kernel function parameters and value of the regularization parameter, which must be set by a user and shouldn't change [1], [2]. It is impossible to provide implementing of high-accuracy data classification with the use of the SVM classifier without adequate solution of this problem.

Let values of the parameters of the SVM classifier be optimal, if high accuracy of classification has been achieved: numbers of error within training and test sets are minimal, moreover the number of errors within test set must not strongly differ from the number of errors within training set. It will allow excluding retraining of the SVM classifier.

In the simplest case solution of this problem can be achieved by a search of the kernel function types, values of the kernel function parameters and value of the regularization parameter that demands significant computational expenses. A herewith for an assessment of classification quality, the indicators of classification accuracy, classification completeness, etc. can be used [3].

In most cases of the development of binary classifiers, it is necessary to work with the complex, multiple extremal, multiple parameter objective function.

Gradient methods are not suitable for search of the optimum of such objective function, but search algorithms of stochastic optimization, such as the genetic algorithm [16]–[18], the artificial bee colony algorithm [19], the particle swarm algorithm [20], [21], etc., have been used. earch of the optimal decision is carried out at once in all space of possible decisions.

The particle swarm algorithm (Particle Swarm Optimization, PSO algorithm), which is based on an idea of possibility to solve the optimization problems using modeling of animals' groups' behavior is the simplest algorithm of evolutionary programming because for its implementation it is necessary to be able to determine only value of the optimized function [20], [21].

The traditional approach to the application of the PSO algorithm consists of the repeated applications of the PSO algorithm for the fixed type of the kernel functions to choose optimal values of the kernel function parameters and value of the regularization parameter with the subsequent choice of the best type of the kernel function and values of the kernel function parameters and value of the regularization parameter corresponding to this kernel function type.

Along with the traditional approach to the application of the PSO algorithm a new approach, that implements the simultaneous search for the best type of the kernel function, values of the kernel function parameters and value of the regularization parameter, is offered [22]. Hereafter, particle swarm algorithms corresponding to traditional and modified approaches will be called as the traditional PSO algorithm and the modified PSO algorithm consequently.

The objective of this paper is to fulfill a comparative analysis of the traditional and modified particle swarm algorithms, applied for the development of the SVM classifier, both on the search time of the optimal parameters of the SVM classifier and the quality of data classification.

The rest of this paper is structured as follows. Section II presents the main stages of the SVM classifier development. Then, Section III details the proposed new approach for solving the problem of the simultaneous search of the kernel function type, values of the kernel function parameters and value of the regularization parameter for the SVM classifier. This approach is based on the application of the modified PSO algorithm. Experimental results comparing the traditional PSO algorithm to the modified PSO algorithm follow in Section IV. Finally, conclusions are drawn in Section V.

## II. THE SMV CLASSIFIER

Let the experimental data set be a set in the form of $\{(z_1,y_1),\ldots,(z_s,y_s)\}$, in which each object $z_i \in Z$ ($i = \overline{1,s}$; $s$ is the number of objects) is assigned to number $y_i \in Y = \{+1; -1\}$ having a value of +1 or −1 depending on the class of object $z_i$. A herewith it is assumed that every object $z_i$ is mapped to $q$-dimensional vector of numerical values of characteristics $z_i = (z_i^1, z_i^2, \ldots, z_i^q)$ (typically normalized by values from the interval [0, 1]) where $z_i^l$ is the numeric value of the $l$-th characteristic for the $i$-th object ($i = \overline{1,s}$, $l = \overline{1,q}$) [22]–[25]. It is necessary with use of special function $\kappa(z_i, z_\tau)$, which is called the kernel, to build the classifier $F: Z \to Y$, which compares to the class with number from the set $Y = \{+1; -1\}$ some object from the set $Z$.

To build «the best» SVM classifier it is necessary to realize numerous repeated training and testing on the different randomly generated training and test sets with following determination of the best SVM classifier in terms of the highest possible classification quality provision. The test set contains the part of data from the experimental data set. The size of the test set must be equal to $1/10 - 1/3$ of the size of the experimental data set. The test set doesn't participate in the control of parameters of the SVM-classifier. This set is used for check of classifier's accuracy. The SVM classifier with satisfactory training and testing results can be used to classify new objects [22].

The separating hyperplane for the objects from the training set can be represented by equation $\langle w, z \rangle + b = 0$, where $w$ is a vector-perpendicular to the separating hyperplane; $b$ is a parameter which corresponds to the shortest distance from the origin of coordinates to the hyperplane; $\langle w, z \rangle$ is a scalar product of vectors $w$ and $z$ [1–3]. The condition $-1 < \langle w, z \rangle + b < 1$ specifies a strip that separates the classes. The wider the strip, the more confidently we can classify objects. The objects closest to the separating hyperplane, are exactly on the bounders of the strip.

In the case of linear separability of classes we can choose a hyperplane so that there is no any object from the training set between them, and then maximize the distance between the hyperplanes (width of the strip) $2/<w,w>$, solving the problem of quadratic optimization [1], [2]:

$$\begin{cases} \langle w, w \rangle \to \min, \\ y_i \cdot (\langle w, z_i \rangle + b) \geq 1, \quad i = \overline{1,S}. \end{cases} \quad (1)$$

The problem of the separating hyperplane building can be reformulated as the dual problem of searching a saddle point of the Lagrange function, which reduces to the problem of quadratic programming, containing only dual variables [1], [2]:

$$\begin{cases} -L(\lambda) = -\sum_{i=1}^{S} \lambda_i + \\ \quad + \frac{1}{2} \cdot \sum_{i=1}^{S} \sum_{\tau=1}^{S} \lambda_i \cdot \lambda_\tau \cdot y_i \cdot y_\tau \cdot \kappa(z_i, z_\tau) \to \min_\lambda, \\ \sum_{i=1}^{S} \lambda_i \cdot y_i = 0, \\ 0 \le \lambda_i \le C, \ i = \overline{1, S}, \end{cases} \quad (2)$$

where $\lambda_i$ is a dual variable; $z_i$ is the object of the training set; $y_i$ is a number (+1 or −1), which characterize the class of the object $z_i$ from the experimental data set; $\kappa(z_i, z_\tau)$ is a kernel function; $C$ is a regularization parameter ($C > 0$); $S$ is a quantity of objects in the experimental data set; $i = \overline{1, S}$.

In training of the SVM classifier it is necessary to determine the kernel function type $\kappa(z_i, z_\tau)$, values of the kernel parameters and value of the regularization parameter $C$, which allows finding a compromise between maximizing of the gap separating the classes and minimizing of the total error. A herewith typically one of the following functions is used as the kernel function $\kappa(z_i, z_\tau)$ [1], [3], [26]:

- linear function: $\kappa(z_i, z_\tau) = <z_i, z_\tau>$;
- polynomial function: $\kappa(z_i, z_\tau) = (<z_i, z_\tau> + 1)^d$;
- radial basis function:

$\kappa(z_i, z_\tau) = exp(-<z_i - z_\tau, z_i - z_\tau>/(2 \cdot \sigma^2))$;

- sigmoid function: $\kappa(z_i, z_\tau) = th(k_2 + k_1 \cdot <z_i, z_\tau>)$,

where $<z_i, z_\tau>$ is a scalar product of vectors $z_i$ and $z_\tau$; $d$ [$d \in N$ (by default $d = 3$)], $\sigma$ [$\sigma > 0$ (by default $\sigma^2 = 1$)], $k_2$ [$k_2 < 0$ (by default $k_2 = -1$)] and $k_1$ [$k_1 > 0$ (by default $k_1 = 1$)] are some of parameters; $th$ is a hyperbolic tangent.

These kernel functions allow dividing the objects from different classes.

As a result of the SVM classifier training the support vectors must be determined. These vectors are closest to the hyperplane separating the classes and contain all information about the classes' separation. The main problem dealing with the training of the SVM classifier, is the lack of recommendations for the choice of value of the regularization parameter, the kernel function type and values of the kernel function parameters, which can provide the high accuracy of objects' classification. This problem can be solved with the use of various optimization algorithms, in particular with the use of the PSO algorithm.

### III. THE MODIFIED PSO ALGORITHM

In the traditional PSO algorithm the $n$-dimensional search space ($n$ is the number of parameters which are subject to optimization) is inhabited by a swarm of $m$ agents-particles (elementary solutions). Position (location) of the $i$-th particle is determined by vector $x_i = (x_i^1, x_i^2, \ldots, x_i^n)$, which defines a set of values of optimization parameters. A herewith these parameters can be presented in an explicit form or even absent in analytical record of the objective function $f(x) = f(x^1, x^2, \ldots, x^n)$ of the optimization algorithm (for example, the optimum is the minimum which must be achieved).

The particles must be situated randomly in the search space during the process of initialization. A herewith each $i$-th particle ($i = \overline{1, m}$) has its own vector of speed $v_i \in R^n$ which influence $i$-th particle ($i = \overline{1, m}$) coordinates' values in every single moment of time corresponding to some iteration of the PSO algorithm.

The coordinates of the $i$-th particle ($i = \overline{1, m}$) in the $n$-dimensional search space uniquely determine the value of the objective function $f(x_i) = f(x_i^1, x_i^2, \ldots, x_i^n)$ which is a certain solution of the optimization problem [20] – [22].

For each position of the $n$-dimensional search space where the $i$-th particle ($i = \overline{1, m}$) was placed, the calculation of value of the objective function $f(x_i)$ is performed. A herewith each $i$-th particle remembers the best value of the objective function found personally as well as the coordinates of the position in the $n$-dimensional space corresponding to the value of the objective function. Moreover each $i$-th particle ($i = \overline{1, m}$) «knows» the best position (in terms of achieving the optimum of the objective function) among all positions that had been «explored» by particles (due to it the immediate exchange of information is replicated by all the particles). At each iteration particles correct their velocity to, on the one hand, move closer to the best position which was found by the particle independently and, on the other hand, to get closer to the position which is the best globally at the current moment. After a number of iterations particles must come close to the best position (globally the best for all iterations). However, it is possible that some particles will stay somewhere in the relatively good local optimum.

Convergence of the PSO algorithm depends on how velocity vector correction is performed. There are different approaches to implementation of velocity vector $v_i$ correction for the $i$-th particle ($i = \overline{1, m}$) [20]. In the classical version of the PSO algorithm correction of each $j$-th coordinate of velocity vector ($j = \overline{1, n}$) of the $i$-th particle ($i = \overline{1, m}$) is made in accordance with formula [20]:

$$v_i^j = v_i^j + \hat{\varphi} \cdot \hat{r} \cdot (\hat{x}_i^j - x_i^j) + \tilde{\varphi} \cdot \tilde{r} \cdot (\tilde{x}^j - x_i^j), \quad (3)$$

where $v_i^j$ is the $j$-th coordinate of velocity vector of the $i$-th particle; $x_i^j$ is the $j$-th coordinate of vector $x_i$, defining the position of the $i$-th particle; $\hat{x}_i^j$ is the $j$-th coordinate of the best position vector found by the $i$-th particle during its existence; $\tilde{x}^j$ is the $j$-th coordinate of the globally best position within the particles swarm in which the objective function has the optimal value; $\hat{r}$ and $\tilde{r}$ are random numbers in interval (0, 1), which introduce an element of randomness in the search process; $\hat{\varphi}$ and $\tilde{\varphi}$ are personal and global coefficients for particle acceleration which are constant and determine behavior and effectiveness of the PSO algorithm in general.

With personal and global acceleration coefficients in (3) random numbers $\hat{r}$ and $\tilde{r}$ must be scaled; a herewith the global acceleration coefficient $\tilde{\varphi}$ operates by the impact of the global best position on the speeds of all particles and the personal acceleration coefficient $\hat{\varphi}$ operates by the impact of the personal best position on the velocity of some particle.

Currently different versions of the traditional PSO algorithm are known. In one of the most known canonical version of the PSO algorithm it is supposed to undertake the normalization of the acceleration coefficients $\hat{\varphi}$ and $\tilde{\varphi}$ to make the convergence of the algorithm not so much dependent on the choice of their values [20].

A herewith correction of each $j$-th coordinate of the velocity vector ($j = \overline{1,n}$) of the $i$-th particle ($i = \overline{1,m}$) is performed in accordance with formula:

$$v_i^j = \chi \cdot [v_i^j + \hat{\varphi} \cdot \hat{r} \cdot (\hat{x}_i^j - x_i^j) + \tilde{\varphi} \cdot \tilde{r} \cdot (\tilde{x}^j - x_i^j)], \quad (4)$$

where $\chi$ is a compression ratio;

$$\chi = 2 \cdot K / |2 - \varphi - \sqrt{\varphi^2 - 4 \cdot \varphi}|; \quad (5)$$

$$\varphi = \hat{\varphi} + \tilde{\varphi} \quad (\varphi > 4); \quad (6)$$

$K$ is the some scaling coefficient, which takes values from the interval (0, 1).

When using formula (4) for correction of velocity vector the convergence of the PSO algorithm is guaranteed and there is no need to control the particle velocity explicitly [20].

Let the correction of velocity vector of the $i$-th particle ($i = \overline{1,m}$) is executed in accordance with one of the formulas (3) or (4). The correction of the $j$-th position coordinate of the $i$-th particle ($i = \overline{1,m}$) can be executed in accordance with the formula:

$$x_i^j = x_i^j + v_i^j. \quad (7)$$

Then for each $i$-th particle ($i = \overline{1,m}$) the new value of the objective function $f(x_i)$ can be calculated and the following check must be perfomed: whether a new position with coordinates vector $x_i$ became the best among all positions in which the $i$-th particle has previously been placed. If new position of the $i$-th particle is recognized to be the best at the current moment the information about it must be stored in a vector $\hat{x}_i$ ($i = \overline{1,m}$).

A herewith value of the objective function $f(x_i)$ for this position must be remembered. Then among all new positions of the swarm particles the check of the globally best position must be carried out. If some new position is recognized as the best globally at the current moment, the information about it must be stored in a vector $\tilde{x}$. A herewith value of the objective function $f(x_i)$ for this position must be remembered.

In the case of the SVM classifier's development with the use of the PSO algorithm the swarm particles can be defined by vectors declaring their position in the search space and corded by the kernel function parameters and the regularization parameter: $(x_i^1, x_i^2, C_i)$, where $i$ is a number of particle ($i = \overline{1,m}$); $x_i^1$, $x_i^2$ are the kernel function parameters of the $i$-th particle, [a herewith parameter $x_i^1$ is equal to the kernel function parameters $d$, $\sigma$ or $k_2$ (depending on the kernel function type which corresponds to a swamp particle); parameter $x_i^2$ is equal to the kernel function parameter $k_1$, if the swamp particle corresponds to the sigmoid type of the kernel function, otherwise this parameter is assumed to be zero]; $C_i$ is the regularization parameter.

Then traditional approach to the application of the PSO algorithm in developing the SVM classifier must be concluded in numerous implementation of the PSO algorithm under the fixed kernel function type aiming to choose the optimal parameters values of the kernel function and value of the regularization parameter.

As result for each type $T$ of the kernel function, participating in the search, the particle with the optimal combination of the parameters values $(\tilde{x}^1, \tilde{x}^2, \tilde{C})$ providing high quality of classification will be defined.

The best type and the best values of the required parameters get out by results of the comparative analysis of the best particles received at realization of the PSO algorithm with the fixed kernel function type.

Along with the traditional approach to the application of the PSO algorithm in the development of the SVM classifier there is a new approach that implements a simultaneous search for the best kernel function type $\tilde{T}$, parameters' values $\tilde{x}^1$ and $\tilde{x}^2$ of the kernel function and value of the regularization parameter $\tilde{C}$. At such approach each $i$-th particle in a swamp ($i = \overline{1,m}$) defined by a vector which describes particle's position in the search space: $(T_i, x_i^1, x_i^2, C_i)$, where $T_i$ is the number of the kernel function type (for example, 1, 2, 3 – for polynomial, radial basis and sigmoid functions accordingly);

parameters $x_i^1$, $x_i^2$, $C_i$ are defined as in the previous case. A herewith it is possible to «regenerate» particle through changing its coordinate $T_i$ on number of that kernel function type, for which particles show the highest quality of classification. In the case of particles' «regeneration» the parameters' values change so that they corresponded to new type of the kernel function (taking into account ranges of change of their values). Particles which didn't undergo «regeneration», carry out the movement in own space of search of some dimension.

The number of particles taking part in «regeneration» must be determined before start of algorithm. This number must be equal to 15% – 25% of the initial number of particles. It will allow particles to investigate the space of search. A herewith they won't be located in it for a long time if their indicators of accuracy are the worst.

The offered modified PSO algorithm can be presented by the following consequence of steps.

Step 1. To determine parameters of the PSO algorithm: number $m$ of particles in a swamp, velocity coefficient $K$, personal and global velocity coefficients $\hat{\varphi}$ and $\tilde{\varphi}$, maximum iterations number $N_{\max}$ of the PSO algorithm. To determine types $T$ of kernel functions, which take part in the search ( $T=1$ – polynomial function, $T=2$ – radial basis function, $T=3$ – sigmoid function) and ranges boundaries of the kernel function parameters and the regularization parameter $C$ for the chosen kernel functions' types $T$: $x_{\min}^{1T}$, $x_{\max}^{1T}$, $x_{\min}^{2T}$, $x_{\max}^{2T}$, $C_{\min}^T$, $C_{\max}^T$ ( $x_{\min}^{2T}=0$ and $x_{\max}^{2T}=0$ for $T=1$ and $T=2$ ). To determine the particles' «regeneration» percentage $p$.

Step 2. To define equal number of particles for each kernel type function $T$, included in search, to initialize coordinate $T_i$ for each $i$-th particle ( $i=\overline{1,m}$ ) (a herewith every kernel function type must be corresponded by equal number of particles), other coordinates of the $i$-th particle ($i=\overline{1,m}$) must be generated randomly from the corresponding ranges: $x_i^1 \in [x_{\min}^{1T},\ x_{\max}^{1T}]$, $x_i^2 \in [x_{\min}^{2T},\ x_{\max}^{2T}]$ ( $x_i^2=0$ under $T=1$ and $T=2$ ), $C_i \in [C_{\min}^T,\ C_{\max}^T]$. To initialize random velocity vector $v_i(v_i^1,v_i^2,v_i^3)$ of the $i$-th particle ( $i=\overline{1,m}$ ) ( $v_i^2=0$ under $T=1$ and $T=2$ ). To establish initial position of the $i$-th particle ( $i=\overline{1,m}$ ) as its best known position $(\hat{T}_i,\hat{x}_i^1,\hat{x}_i^2,\hat{C}_i)$, to determine the best particle with coordinates' vector $(\tilde{T},\tilde{x}^1,\tilde{x}^2,\tilde{C})$ from all the $m$ particles, and to determine the best particle for each kernel function type $T$, including in a search, with coordinates' vector $(\overline{T},\overline{x}^{1T},\overline{x}^{2T},\overline{C}^T)$. Herewith number of executed iterations must be considered as 1.

Step 3. To execute while the number of iterations is less than the fixed number $N_{\max}$:

- «regeneration» of particles: to choose $p$ % of particles which represent the lowest quality of classification from particles with coordinate $T_i \neq \tilde{T}$ ( $i=\overline{1,m}$ ); to change coordinate $T_i$ (with the kernel function type) on $\tilde{T}$; to change values of the parameters $x_i^1,x_i^2,C_i$ of «regenerated» particles to let them correspond to a new kernel function type $\tilde{T}$ (within the scope of the corresponding ranges);

- correction of velocity vector $v_i(v_i^1,v_i^2,v_i^3)$ and position $(x_i^1,x_i^2,C_i)$ of the $i$-th particle ( $i=\overline{1,m}$ ) using formulas:

$$v_i^j = \begin{cases} \chi \cdot [v_i^j + \hat{\varphi}\cdot\hat{r}\cdot(\hat{x}_i^j - x_i^j) + \tilde{\varphi}\cdot\tilde{r}\cdot(\overline{x}^{jT} - x_i^j)], & j=1,\ 2, \\ \chi \cdot [v_i^j + \hat{\varphi}\cdot\hat{r}\cdot(\hat{C}_i - C_i) + \tilde{\varphi}\cdot\tilde{r}\cdot(\overline{C}^T - C_i)], & j=3, \end{cases}$$
(8)

$$x_i^j = x_i^j + v_i^j \text{ for } j=1,\ 2, \quad (9)$$

$$C_i = C_i + v_i^3, \quad (10)$$

where $\hat{r}$ and $\tilde{r}$ are random numbers in interval (0, 1), $\chi$ is a compression ratio calculated using the formula (5); a herewith formula (8) is the modification of formula (4): the coordinates' values $\overline{x}^{1T},\overline{x}^{2T},\overline{C}^T$ are used instead of the coordinates' values $\tilde{x}^1,\tilde{x}^2,\tilde{C}$ of the globally best particle;

- accuracy calculation of the SVM classifier with parameters' values $(T_i,x_i^1,x_i^2,C_i)$ ( $i=\overline{1,m}$ ) with aim to find the optimal combination $(\tilde{T},\tilde{x}^1,\tilde{x}^2,\tilde{C})$, which will provide high quality of classification;

- increase of iterations number on 1.

The particle with the optimal combination of the parameters' values $(\tilde{T},\tilde{x}^1,\tilde{x}^2,\tilde{C})$ which provides the highest quality of classification on chosen the function types will be defined after execution of the offered algorithm.

After executing of the modified PSO algorithm it can be found out that all particles will be situated in the search space which corresponds to the kernel function with the highest classification quality because some particles in the modified PSO algorithm changed their coordinate, which is responsible for number of the kernel function. A herewith all other search spaces will turn out to be empty because all particles will «regenerate» their coordinate with number of the kernel function type. In some cases (for small values of the iterations' number $N_{\max}$ and for small value of the particles' «regeneration» percentage $p$ ) some particles will not «regenerate» their kernel function type and will stay in their initial search space.

Using of this approach in the application of the PSO algorithm in the problem of the SVM classifier development allows reducing the time required to construct the desired SVM classifier.

Quality evaluation of the SVM classifier can be executed with the use of different classification quality indicators [3]. There are cross validation data indicator, accuracy indicator, classification completeness indicator and ROC curve analysis based indicator, etc.

## IV. EXPERIMENTAL STUDIES

The feasibility of the modified PSO algorithm using for the SVM classifier development was approved by test and real data. In the experiment for a particular data set the traditional PSO algorithm and the modified PSO algorithm were carried out. Comparison between these algorithms was executed using the found optimal parameters values of the SVM algorithm, classification accuracy and spent time.

Actual data used in the experimental researches was taken from Statlog project and from UCI machine learning library. Particularly, we used two data sets for medical diagnostics and one data set for credit scoring:

- breast cancer data set of The Department of Surgery at the University of Wisconsin, in which the total number of instances is 569 including 212 cases with the diagnosed cancer (class 1) and 357 cases without such diagnosis (class 2); a herewith each patient is described by 30 characteristics ($q = 30$) and all information was obtained with the use of digital images (WDBC data set in the Table, the source is http://archive.ics.uci.edu/ml/; machine-learning-databases/breast-cancer-wisconsin/);

- heart disease data set, in which the total number of instances is 270 including 150 cases with the diagnosed heart disease (class 1) and 120 cases without such diagnosis (class 2); a herewith each patient is described by 13 characteristics ($q = 13$) (Heart data set in the Table, the source is http://archive.ics.uci.edu/ml/machine-learning-databases/statlog/heart/; a herewith desease was found for 150 patients (class 1) and desease was not found for 120 patients (class 2));

- Australian consumer credit data set, in which the total number of instances is 690 including 382 creditworthy cases (class 1) and 308 default cases (class 2); a herewith each applicant is described by 14 characteristics ($q = 14$) (Australian data set in the Table, the source is http://archive.ics.uci.edu/ml/machine-learning-data bases/statlog/australian/).

Moreover two testing data sets were used in experimental researches: Test [11] and МОТП12 (the source is http://machinelearning.ru/wiki/images/b/b2/MOTP12_svm_example.rar).

For all data sets binary classification was performed.

For development of the SVM classifier the traditional and the modified PSO algorithms were used; a herewith the choice of the optimal values of the SVM classifier parameters was realized. The kernels with polynomial, radial basis and sigmoid functions were included in the search and the identical values of the PSO algorithm parameters and the identical ranges of values' change of the required SVM classifier parameters were established.

The short description of characteristics of each data set is provided in the Table. Here search results of the optimal values of parameters of the SVM classifier with the application of the traditional PSO algorithm and the modified PSO algorithm are presented (in the identical ranges of parameters' change and at the identical PSO algorithm parameters), number of error made during the training and testing of the SVM classifier and search time. For example, for WDBC data set with the use of the traditional and the modified PSO algorithms the kernel with radial basis function (number 2) was determined as the optimal. For the traditional PSO algorithm the optimal values of the kernel parameter and the regularization parameter are equal to $\sigma = 6.81$ and $C = 4.93$ accordingly. For the modified PSO algorithm the optimal values of the kernel parameter and the regularization parameter are equal to $\sigma = 4.01$ and $C = 9.83$ accordingly.

The classification accuracy by the traditional PSO algorithm is equal to 99.12%, and the classification accuracy by the modified PSO algorithm is equal to 99.65%. A herewith the search time came to 10108 and 3250 seconds accordingly.

For Heart data set in the Figures 2 – 4 the examples of position of the particles swarm in the D-2 search spaces and in the D-3 search space during initialization, at the 3-rd iteration and at the 12-th iteration (with the use of the modified PSO algorithm) are shown.

The kernels with polynomial, radial basis and sigmoid functions were included in the search. A herewith the following change ranges of values' parameters were set: $3 \leq d \leq 8$, $d \in N$ (for polynomial function); $0.1 \leq \sigma \leq 10$ (for radial basis function); $-10 \leq k_2 \leq -0.1$ и $0.1 \leq k_1 \leq 10$ (for sigmoid function).

TABLE I. THE SEARCH RESULTS BY MEANS OF THE TRADITIONAL PSO ALGORITHM AND THE MODIFIED PSO ALGORITHM

| Data set | Number of objects | Number of characteristics | PSO algorithm type | Found parameters | | | | Errors | | Number of support vectors | Accuracy (%) | Search time (sec.) |
|---|---|---|---|---|---|---|---|---|---|---|---|---|
| | | | | Kernel number | $C$ | $x_1$ | $x_2$ | During the training | During the testing | | | |
| WDBC | 569 | 30 | traditional | 2 | 4.93 | 6.81 | - | 5 of 427 | 0 of 142 | 66 | 99.12 | 10108 |
| | | | modified | 2 | 9.83 | 4.01 | - | 1 of 427 | 1 of 142 | 80 | 99.65 | 3250 |
| Heart | 270 | 13 | traditional | 2 | 8.51 | 3.18 | - | 7 of 192 | 11 of 78 | 106 | 93.33 | 6558 |
| | | | modified | 2 | 5.92 | 2.89 | - | 4 of 192 | 10 of 78 | 99 | 94.81 | 2733 |
| Australian | 690 | 14 | traditional | 2 | 6.79 | 2.31 | - | 21 of 492 | 28 of 198 | 237 | 92.9 | 16018 |
| | | | modified | 2 | 7.64 | 2.38 | - | 21 of 492 | 26 of 198 | 225 | 93.19 | 8013 |
| MOTII12 | 400 | 2 | traditional | 2 | 8.10 | 0.18 | - | 9 of 340 | 9 of 60 | 166 | 95.50 | 15697 |
| | | | modified | 2 | 6.37 | 0.26 | - | 10 of 340 | 8 of 60 | 107 | 95.50 | 9145 |
| Test | 300 | 2 | traditional | 1 | 8.85 | 3 | - | 0 of 240 | 0 of 60 | 7 | 100 | 3433 |
| | | | modified | 1 | 9.24 | 3 | - | 0 of 240 | 0 of 60 | 6 | 100 | 648 |

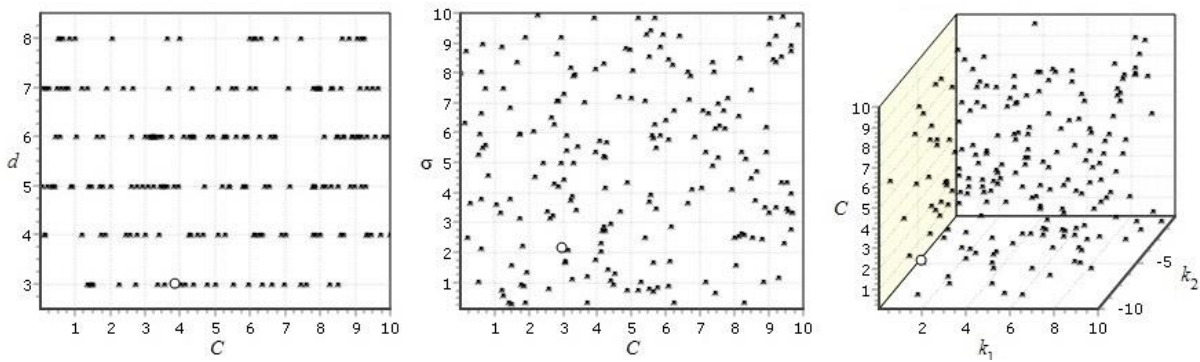

Fig. 2. Location of particles in a swamp during the initialization (polynomial kernel function is on the left, radial basis is in the middle, sigmoid is on the right)

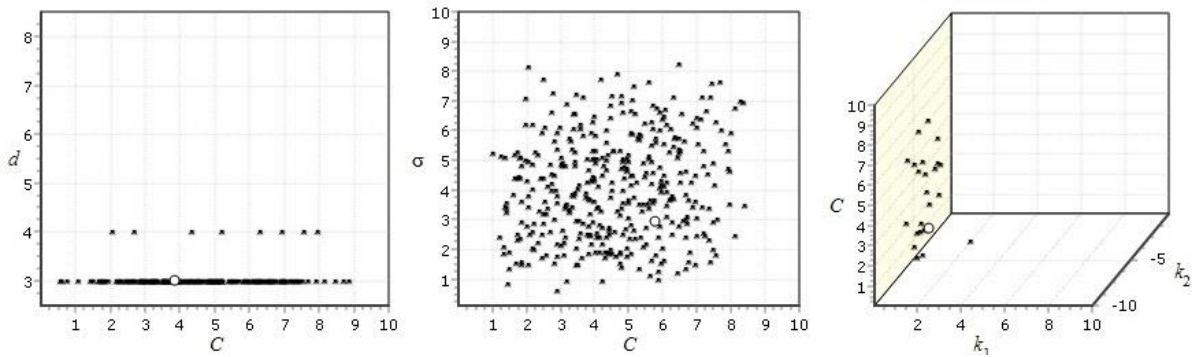

Fig. 3. Location of particles in a swamp during the 3-rd iteration (polynomial kernel function is on the left, radial basis is in the middle, sigmoid is on the right)

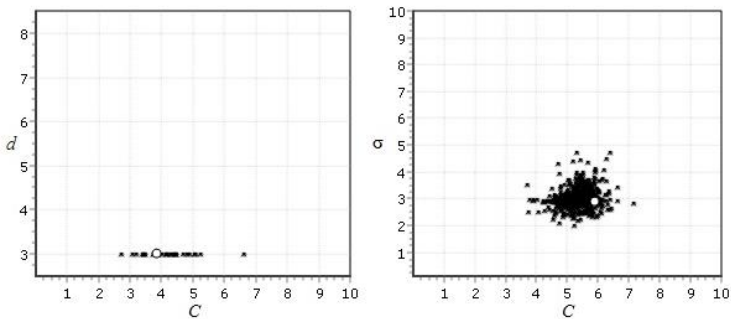

Fig. 4. Location of particles in a swamp during the 12-th iteration (radial basis kernel function is on the left and sigmoid kernel function is on the right)

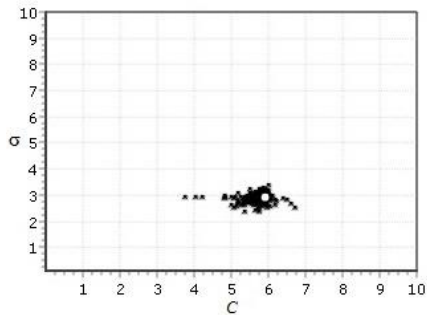

Fig. 5. Location of particles after the 20-th iteration

Change range for the regularization parameter $C$ was determined as: $0.1 \leq C \leq 10$. Moreover, the following values of parameters of the PSO algorithm were set: number $m$ of particles in a swarm equal to 600 (200 per each kernel function type); iterations' number $N_{max} = 20$; personal and global velocity coefficients equal to $\hat{\varphi} = 2$ and $\tilde{\varphi} = 5$ accordingly; the scaling coefficient $K = 0.3$; «regeneration» coefficient of particles $p = 20\%$. Particles marked by asterisk bullets in the search spaces and the best position from the search space is marked by white round bullet. During realization of the modified PSO algorithm the swamp particles moves towards the best (optimal) position for the current iteration in the search space and demonstrate collective search of the optimal position. A herewith velocity and direction of each particle are corrected. Moreover «regeneration» of particles takes place: some particles change own search space to space, in which particles show the best quality of classification.

Thus, during realization of the modified PSO algorithm there is a change of the particles' coordinates, which are responsible for parameters of the kernel function $\kappa(z_i, z_\tau)$ and the regularization parameter $C$. Besides, the type of the kernel function also changes. As a result the particles moves towards the united search space (in this case – the space corresponding to radial basis kernel function) leaving the space where they were initialized.

In the reviewed example only 7 particles didn't change their kernel function type after 20 iterations. Other particles situated near the best position responsible for the optimal solution in the search space (Figure 5).

It is visible from the Table that that as a result of search for the reviewed data sets both algorithms determined identical kernel function type as the optimal, similar values of the kernel function parameter and the regularization parameter, and also similar accuracy values of training and testing of the SVM classifier.

But the modified PSO algorithm is more effective, because it took less (more than 2–3 times) time for search than traditional one.

## V. CONCLUSION

The experimental results obtained on the base of the test data traditionally used to assess the classification quality, confirm the efficiency of the modified PSO algorithm. This algorithm allows choosing the best kernel function type, values of the kernel function parameters and value of the regularization parameter with the time expenditures which are significantly less, than in the case of the traditional PSO algorithm. A herewith high accuracy of classification is provided.

The obtained results had been reached thanks to «regeneration» of particles in the modified PSO algorithm. Particles which participate in the «regeneration» process change their kernel function type to the one which corresponds to the particle with the best value of the classification accuracy. Also, these articles change the accessory ranges of their parameters.

Further researches will have been devoted to the development of recommendations on the application of the modified PSO algorithm in the solution of the practical problems.

REFERENCES

[1] O. Chapelle, V. Vapnik, O. Bousquet, and S. Mukherjee, "Choosing Multiple Parameters for Support Vector Machine," Machine Learning, vol. 46, no. 1–3, pp. 131–159, 2002.
[2] V. Vapnik, Statistical Learning Theory. Wiley, New York, 1998.
[3] L. Yu, S. Wang, K. K. Lai and L. Zhou, Bio-Inspired Credit Risk Analysis. Computational Intelligence with Support Vector Machines. Springer-Verlag, 2008.
[4] J.S. Raikwal and K. Saxena, "Performance Evaluation of SVM and K-Nearest Neighbor Algorithm over Medical Data set," International Journal of Computer Applications, vol. 50, no. 14, pp. 35–39, 2012.
[5] Y. LeCun, L.D. Jackel, L. Bottou, C. Cortes at al., "Learning Algorithms for Classification: A Comparison on Handwritten Digit Recognition," in Neural Networks: The Statistical Mechanics Perspective, J. H. Oh, C. Kwon and S. Cho, Eds. World Scientific, 1995, pp. 261–276.
[6] T. Joachims, "Text Categorization with Support Vector Machines: Learning with Many Relevant Features," Lecture Notes in Computer Science, vol. 1398, pp. 137–142, 2005.
[7] Y. Li, K. Bontcheva and H. Cunningham, "SVM Based Learning System For Information Extraction," Lecture Notes in Computer Science, vol. 3635, pp. 319–339, 2005.
[8] M. Oren, C. Papageorgious, P. Sinha, E. Osuna, and T. Poggio, "Pedestrian Detection Using Wavelet Templates," in 1997 IEEE Computer Society Conference on Computer Vision and Pattern Recognition, 1997, pp. 193–199.
[9] E. Osuna, R. Freund and F. Girosi, "Training Support Vector Machines: An Application to Face Detection," in 1997 IEEE Computer Society Conf. on Computer Vision and Pattern Recognition, 1997, pp. 130–136.
[10] J.C. Platt, "Fast Training of Support Vector Machines Using Sequential Minimal Optimization," in Advances in Kernel Methods. Support Vector Learning, 1998, pp. 185–208.
[11] S.K. Shevade, S.S. Keerthi, C. Bhattacharyya and K.R.K. Murthy, "Improvements to the SMO Algorithm for SVM Regression," IEEE Trans. on Neural Networks, vol. 11, no. 5, pp. 1188–1193, 2000.
[12] E. Osuna, R. Freund and F. Girosi, "Improved Training Algorithm for Support Vector Machines," in 1997 IEEE Workshop Neural Networks for Signal Processing, 1997, pp. 24–26.

[13] S.V.N. Vishwanathan, A. Smola and N. Murty, "SSVM: a simple SVM algorithm," Proceedings of the 2002 International Joint Conference on Neural Networks, vol. 3, pp. 2393-2398, 2002.

[14] S. Shalev-Shwartz, Y. Singer, N. Srebro and A. Cotter, "Pegasos: Primal Estimated sub-Gradient Solver for SVM," Mathematical Programming, vol. 127, no. 1, pp. 3–30, 2011.

[15] L. Bottou and C.-J. Lin. Support Vector Machine Solvers, 2007.

[16] D.E. Goldberg, B. Korb and K. Deb, "Messy genetic algorithms: Motivation, analysis, and first results," Complex Systems, vol. 3, no. 5, pp. 493–530, 1989.

[17] D.R. Eads, D. Hill, S. Davis, S.J. Perkins, J. Ma at al., "Genetic algorithms and support vector machines for time series classification," in Proc. SPIE 4787 Applications and Science of Neural Networks, Fuzzy Systems, and Evolutionary Computation, vol. 74, 2002, p. 74.

[18] S. Lessmann, R. Stahlbock and S.F. Crone, "Genetic algorithms for support vector machine model selection," in 2006 International Joint Conference on Neural Networks, 2006, pp. 3063–3069.

[19] D. Karaboga and B. Basturk, "Artificial Bee Colony (ABC) Optimization Algorithm for Solving Constrained Optimization Problems," in Proc. of the 12th Intern. Fuzzy Systems Association world congress on Foundations of Fuzzy Logic and Soft Computing, 2007, pp. 789–798.

[20] J. Sun, C.-H. Lai and X.-J. Wu, Particle Swarm Optimisation: Classical and Quantum Perspectives. CRC Press, 2011.

[21] R. Poli, J. Kennedy and T. Blackwell, "Particle swarm optimization," Swarm Intelligence, vol. 1, no. 1, pp. 33–57, 2007.

[22] L. Demidova and Yu. Sokolova, "Modification Of Particle Swarm Algorithm For The Problem Of The SVM Classifier Development," in 2015 International Conference "Stability and Control Processes" (SCP), 2015, pp. 623–627.

[23] L. Demidova, Yu. Sokolova and E. Nikulchev, "Use of Fuzzy Clustering Algorithms' Ensemble for SVM classifier Development," International Review on Modelling and Simulations, vol. 8, no. 4, pp. 446–457, 2015.

[24] L. Demidova and Yu. Sokolova, "SVM-Classifier Development With Use Of Fuzzy Clustering Algorithms' Ensemble On The Base Of Clusters' Tags' Vectors' Similarity Matrixes," in 16th International Symposium on Advanced Intelligent Systems, 2015, pp. 889–906.

[25] L. Demidova, and Yu. Sokolova, "Training Set Forming For SVM Algorithm With Use Of The Fuzzy Clustering Algorithms Ensemble On Base Of Cluster Tags Vectors Similarity Matrices," in 2015 International Conference Stability and Control Processes (SCP), pp. 619–622, 2015.

[26] A. Karatzoglou, D. Meyer and K. Hornik, Support vector machines in R. Research Report, WU Vienna, 2005.